\pdfoutput=1

\documentclass[11pt]{article}

\usepackage[]{acl}

\usepackage{times}
\usepackage{latexsym}

\usepackage[T1]{fontenc}

\usepackage[utf8]{inputenc}

\usepackage{algorithm}
\usepackage{microtype}
\usepackage{algcompatible,amsmath}
\usepackage{graphicx}
\usepackage{url}
\usepackage{multirow}
\usepackage{comment}
\usepackage{amssymb}
\usepackage{pifont}
\usepackage{balance}
\usepackage{mathtools}
\usepackage{xcolor}
\usepackage{colortbl}
\usepackage{bbm}
\usepackage{enumitem}
\usepackage{marvosym}
\usepackage{textcomp}
\usepackage{array}
\usepackage{booktabs}
\usepackage{tabularx}
\usepackage{CJKutf8}
\usepackage{algorithmicx}  
\usepackage{algpseudocode} 
\usepackage{makecell}

\usepackage{newfloat}
\usepackage{listings}
\floatstyle{ruled}
\newfloat{listing}{tb}{lst}{}
\floatname{listing}{Listing}

\newcommand{\squishlist}{
\begin{list}{$\bullet$}
{ \usecounter{Lcount}
\setlength{\itemsep}{0pt}
\setlength{\parsep}{0pt}
\setlength{\topsep}{0pt}
\setlength{\partopsep}{0pt}
\setlength{\leftmargin}{2em}
\setlength{\labelwidth}{1.5em}
\setlength{\labelsep}{0.5em} } }
\newcommand{\squishend}{
\end{list} }
\makeatletter
\newcommand{\thickhline}{%
    \noalign {\ifnum 0=`}\fi \hrule height 2pt
    \futurelet \reserved@a \@xhline
}

\title{Multimodal Relation Extraction with Cross-Modal Retrieval and Synthesis}

\author{Xuming Hu$^{1}$, Zhijiang Guo$^{2\dagger}$, Zhiyang Teng$^{3}$, Irwin King$^4$, Philip S. Yu$^{1,5}$\\
  $^1$Tsinghua University, $^2$University of Cambridge, $^3$Nanyang Technological University,\\ 
  $^4$The Chinese University of Hong Kong, $^5$University of Illinois at Chicago\\
  $^1$\texttt{hxm19@mails.tsinghua.edu.cn}
  $^2$\texttt{zg283@cam.ac.uk}\\
  $^3$\texttt{zhiyang.teng@ntu.edu.sg}  $^4$\texttt{king@cse.cuhk.edu.hk}  $^5$\texttt{psyu@uic.edu}
  }
\begin{document}
\maketitle
\begin{abstract}

Multimodal relation extraction (MRE) is the task of identifying the semantic relationships between two entities based on the context of the sentence image pair. Existing retrieval-augmented approaches mainly focused on modeling the retrieved textual knowledge, but this may not be able to accurately identify complex relations. To improve the prediction, this research proposes to retrieve textual and visual evidence based on the object, sentence, and whole image. We further develop a novel approach to synthesize the object-level, image-level, and sentence-level information for better reasoning between the same and different modalities. Extensive experiments and analyses show that the proposed method is able to effectively select and compare evidence across modalities and significantly outperforms state-of-the-art models. Code and data are available\footnote{\url{https://github.com/THU-BPM/MRE}\\\phantom{00} $^\dagger$Corresponding Author.}.

\end{abstract}
\section{Introduction}
\label{sec:intro}

\begin{figure}
    \centering
    \includegraphics[scale=1.3]{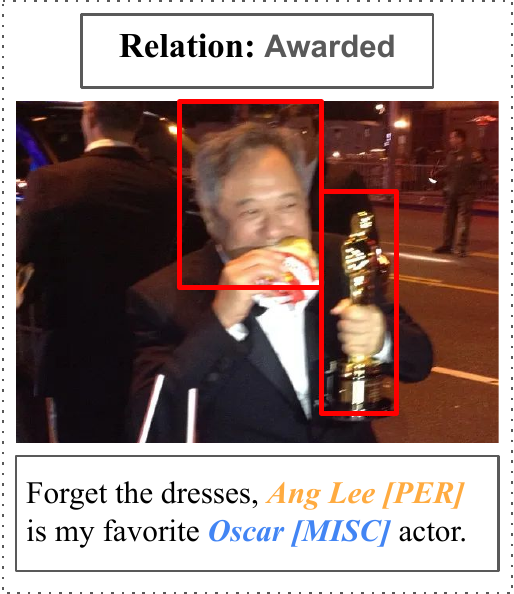}
    \caption{Example from MNRE. Entities are highlighted. Objects are denoted by the bounding boxes. }
    \label{fig:example}
\end{figure}

Relation extraction aims to detect relations among entities in the text and plays an important role in various applications~\citep{ZhangZCAM17,soares2019matching}. Early efforts mainly focus on predicting the relations based on the information from one single modality i.e., text. Recently, multimodal relation extraction (MRE) has been proposed to enhance textual representations with the aid of visual clues from images~\citep{ZhengFFCL021,Chen2022,Wang2022NamedEA}. It extends the text-based approaches by providing visual contexts to address the common ambiguity issues in identifying relations. Figure~\ref{fig:example} shows an example from the MNRE dataset~\citep{ZhengWFF021}.  To infer the relation between entities \textit{Ang Lee} and \textit{Oscar}, the model needs to capture the interactions from visual relations between objects in an image to textual relations in a sentence.  The visual relation ``holding'' between two objects helps to detect the relation \textbf{awarded} between two textual entities.


Most existing efforts focus on modeling the visual and textual content of the input. \citet{ZhengFFCL021} constructed the textual and visual graphs, then identify the relations based on graph alignments. \citet{Chen2022} presents a hierarchical visual prefix fusion network to incorporate hierarchical multi-scaled visual and textual features. \citet{li2022analyzing} proposes a fine-grained multimodal alignment approach with Transformer, which aligns visual and textual objects in representation space. \citet{Wang2022NamedEA} first proposes retrieval-augmented multimodal relation extraction. The given image and sentence are used to retrieve textual evidence from the knowledge base constructed based on Wikipedia. Unlike previous retrieval-based models, we not only retrieve texts but also retrieve visual and textual evidence related to the object, sentence, and entire image. A novel strategy is used to combine evidence from the object, sentence, and image levels in order to make better reasoning across modalities. Our key contributions are summarized as follows:


\begin{itemize}
    \item We use cross-modal retrieval for obtaining multimodal evidence. To improve prediction accuracy, we further synthesize visual and textual information for relational reasoning.
    \item We evaluate our method on the MRE benchmark. Extensive experimental results validate the effectiveness of the proposed approach.
\end{itemize}



\section{Methodology}
\label{sec:baseline}

\begin{figure}
    \centering
    \includegraphics[scale=0.32]{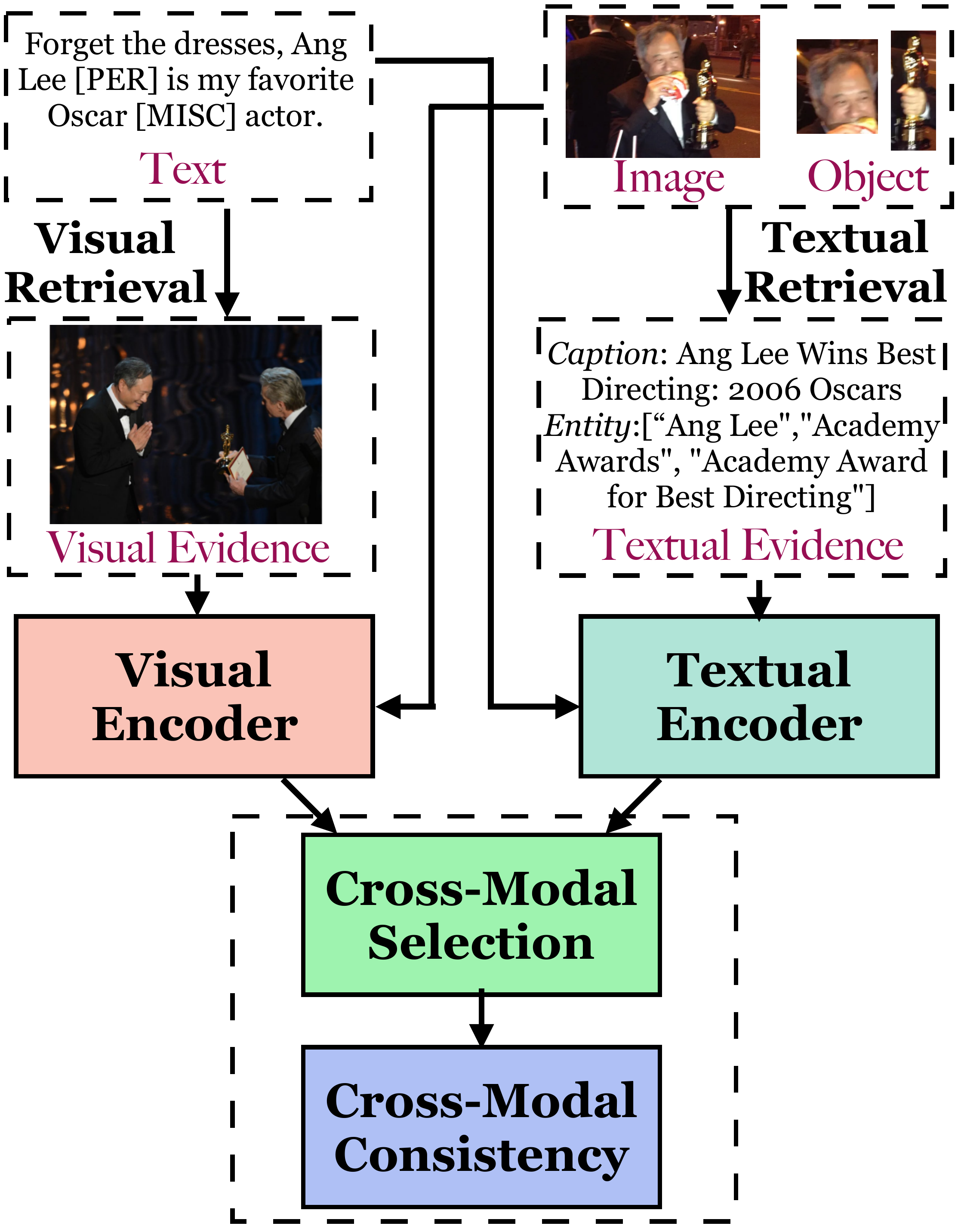}
    \caption{Overview of the model.}
    \label{fig:arch}
\end{figure}


\subsection{Cross-Modal Retrieval}
This module aims to retrieve visual evidence based on the input text (sentence, entities), and textual evidence based on the input image and objects. 

\paragraph{Textual evidence} 
We first obtain the local visual objects with top $m$ salience by using the visual grounding toolkit~\citep{yang2019fast}: $V_{obj}=\{V_{obj}^{1}, V_{obj}^{2},\cdots,V_{obj}^{m}\}$. Then we retrieve $V_{img}$ and $V_{obj}$ using Google Vision APIs\footnote{\url{https://cloud.google.com/vision/docs/detecting-web}} to obtain textual evidence, which returns a list of entities $E_{entity}$ that describe the content of the $V_{img}$ and $V_{obj}$ and provide a more effective explanation for the visual content. In addition to $Entity$, the APIs could return the images’ URLs and the containing pages’ URLs. We propose a web crawler to search the images’ URLs in the containing pages’ and then return the captions $E_{caption}$ if found. Note that $E_{entity}$ and $E_{caption}$ contain 10 entities and captions obtained for each $V_{img}$ and $V_{obj}$ as retrieval textual evidence.

%

\paragraph{Visual Evidence} 
We use the textual content $T$ of the post to retrieve the visual evidence. More specially, we leverage the Google custom search API\footnote{\url{https://developers.google.com/custom-search/v1}} to retrieve the 10 images $E_{image}$ for the textual content in each post.
            

\subsection{Cross-Modal Synthesis}

Given the retrieved visual and textual evidence, this module aims to synthesize multimodal information for relation extraction.






\subsubsection{Visual Encoder}

The visual encoder module encodes the visual content $V_{img}$, $V_{obj}$ and retrieved visual evidence $E_{image}$ of the post. First, we adopt the ResNet~\citep{HeZRS16} which is pretrained on the ImageNet dataset~\citep{DengDSLL009} to obtain the visual embedding $\boldsymbol{h}_{v}\in \mathbb{R}^{n\times d}$, where $n$ and $d$ represents the number of images and the hidden dimension. To fuse the cross-modal visual and textual information, we employ a learnable linear layer $\boldsymbol{h}_{v}=\mathbf{W}_{\phi}\boldsymbol{h}_{v}+\mathbf{b}_{\phi}$

\subsubsection{Textual Encoder}

The textual module encodes the textual content $T$ and retrieved textual evidence $E_{entity}$, $E_{caption}$ of the post. For each sentence ${X = [x_{1}, x_{2},.., x_{M}]}$ in the textual content $T$ where two entities ${[E_{1}]}$ and ${[E_{2}]}$ are mentioned, we follow the labeling schema adopted in \citet{soares2019matching} and argument ${X}$ with four reserved tokens ${[E_{1}]}$, ${[/E_{1}]}$, ${[E_{2}]}$, ${[/E_{2}]}$ to mark the beginning and the end of each entity mentioned in the sentence:
\begin{equation}
\begin{split}
{X}=&\big[x_{1},...,[E_1],x_{i},...,x_{j-1},[/E_1],\\
\qquad...,&[E_2],x_{k},...,x_{l-1},[/E_2],...,x_{M}\big],
\end{split}
\end{equation}
as the input token sequence. We adopt BERT~\citep{Devlin2019BERTPO} as an encoder and obtain the textual embedding $\boldsymbol{h}_{t}\in \mathbb{R}^{(M+4)\times d}$, where $M$ and $d$ represents the number of tokens in $s$ and the hidden dimensions. Thanks to informative visual embeddings, we can better capture the correlation between visual content and textual information.




\subsubsection{Cross-Modal Selection}

Given the encoded multimodal evidence and inputs $\boldsymbol{h}_{t}^{l}\in \mathbb{R}^{(M+4)\times d}$,  $\boldsymbol{h}_{v}^{l}\in \mathbb{R}^{n\times d}$. The module selects visual/textual evidence and compares it against the input image/sentence.  Inspired by~\citet{Vaswani2017AttentionIA}, we leverage multi-head attention to perform the cross-modal selection. We first project the presentations as query, key, and value vectors:


\begin{small}
\begin{equation}\label{attention_1}
    \begin{split}
        \boldsymbol{Q}^l, \boldsymbol{K}^l, \boldsymbol{V}^l=\boldsymbol{x} \boldsymbol{W}_q^l, \boldsymbol{x} \boldsymbol{W}_k^l, \boldsymbol{x} \boldsymbol{W}_v^l ; \boldsymbol{x} \in\left\{\boldsymbol{h}_t^l, \boldsymbol{h}_v^l\right\},
    \end{split}
\end{equation}
\end{small}
where $\boldsymbol{W}_q^l, \boldsymbol{W}_k^l, \boldsymbol{W}_v^l \in \mathbb{R}^{d \times d_h}$ represent attention projection parameters. We then obtain the hidden features at ($l+1$)-th layer via multi-head attention:
\begin{equation}\label{attention_2}
\begin{split}
& \boldsymbol{h}_t^{l+1}=\operatorname{Attn}\left(\boldsymbol{Q}_t^l,\left[\boldsymbol{K}_v^l, \boldsymbol{K}_t^l\right],\left[\boldsymbol{V}_v^l, \boldsymbol{V}_t^l\right]\right), \\
& \boldsymbol{h}_v^{l+1}=\operatorname{Attn}\left(\boldsymbol{Q}_v^l,\left[\boldsymbol{K}_t^l, \boldsymbol{K}_v^l\right],\left[\boldsymbol{V}_t^l, \boldsymbol{V}_v^l\right]\right).
\end{split}
\end{equation}
Note that the textual features $\boldsymbol{h}_{t}$ come from two types: The first is the textual content in the post with two entities, so we get the relational features of the ${[E_{1}]}$and ${[E_{2}]}$ positions. The other is retrieved textual evidence, since it does not have entities, we obtain representations of the $CLS$ position:
\begin{equation}\label{attention_3}
\begin{split}
\boldsymbol{h}_{t,content} = &\operatorname{Avg.}(\boldsymbol{h}_{t,[E_1]}, \boldsymbol{h}_{t,[E_2]}),\\
\boldsymbol{h}_{t,retrieved} &= \boldsymbol{h}_{t,[CLS]}.
\end{split}
\end{equation}
where $\boldsymbol{h}_{t}=\{\boldsymbol{h}_{t,content}, \mathbf{h}_{t,retrieved}\}  \in \mathbb{R}^{d}$ is the representation of the textual content and retrieved textual evidence for each post, where $d$ is the embedding size $768$. Similarly, we use a learnable linear layer $\boldsymbol{h}_{t}=\mathbf{W}_{\theta}\boldsymbol{h}_{t}+\mathbf{b}_{\theta}$ to change the dimension $d$ from $768$ to $2048$ and employ the multi-head attention in Eq. \ref{attention_1}, \ref{attention_2}, and \ref{attention_3} to update the visual content and retrieved visual evidence.




\subsubsection{Cross-Modal Consistency}
This module aims to evaluate the consistency between the retrieved textual and visual evidence and the original post. A natural idea is to leverage the textual and visual content in the original post to update the retrieved textual and visual evidence. We could obtain the updated evidence $\boldsymbol{h}_{t,retrieved}$ and $\boldsymbol{h}_{v,retrieved}$ with $\boldsymbol{h}_{t,content}$ and $\boldsymbol{h}_{v,content}$ as:
\begin{equation}
\begin{split}
\boldsymbol{h}_{t,r.} = \operatorname{softmax}(\frac{\boldsymbol{h}_{t,c.}\mathbf{W}_{t} \times (\boldsymbol{h}_{t,r.}\mathbf{W}_{t}^{'})^{T}}{\sqrt{d_{t}}})\boldsymbol{h}_{t,r.},\\
\boldsymbol{h}_{v,r.} = \operatorname{softmax}(\frac{\boldsymbol{h}_{t,c.}\mathbf{W}_{v} \times (\boldsymbol{h}_{v,r.}\mathbf{W}_{v}^{'})^{T}}{\sqrt{d_{v}}})\boldsymbol{h}_{v,r.},
\end{split}
\end{equation}
where $\mathbf{W}_{t}, \mathbf{W}_{t}^{'} \in \mathbb{R}^{768 \times 768}$ and $\mathbf{W}_{v}, \mathbf{W}_{v}^{'} \in \mathbb{R}^{2048 \times 2048}$ are trainable projection matrices and $d_{t},d_{v}$ are hyperparameters.

\subsection{Classifier}



We concatenate the resulting representations to form the final multimodal representations and leverage a feed-forward neural network to predict the relation:
\begin{equation}
\boldsymbol h_{final} = \text{FFNN}([\boldsymbol{h}_{t,c.};\boldsymbol{h}_{t,r.};\boldsymbol{h}_{v,c.};\boldsymbol{h}_{v,r.}] ),
\end{equation}
where $\boldsymbol{h}_{final}$ is then fed into a linear layer followed by a softmax operation to obtain a probability distribution $p \in \mathbb{R}^m$ over $m$ relation labels. 


\section{Experiments and Analyses}
\label{sec:experiments}

\subsection{Experimental Setup}
We evaluate the model on MNRE~\citep{ZhengWFF021}, which contains 12,247/1,624/1,614 samples in train/dev/test sets, 9,201 images, and 23 relation types. Following prior efforts, we adopt Accuracy, Precision, Recall, and F1 as the evaluation metrics. For fair comparisons, all baselines and our method use ResNet50~\cite{HeZRS16} as the visual backbone and BERT-base \cite{Devlin2019BERTPO} as the textual encoder. We computed the Accuracy and Macro F1 as the evaluation metric. The hyper-parameters are chosen based on the development set. Results are reported with mean and standard deviation based on 5 runs. For the textual encoder of the retrieval-based model, we use the BERT-Base default tokenizer with a max-length of 128 to preprocess data. For the visual encoder of the retrieval-based model, we use ResNet 50 to encode the visual images. We scale the image proportionally so that the short side is 256, and crop the center to $224*224$. For the feed-forward neural network of the classifier, we set the layer dimensions as ${{h_{R}}}$-${1024}$-verification\_labels, where ${{h_{R}}=768*2+2048*2}$. We use BertAdam with 3e-5 learning rate, warmup with 0.06 to optimize the cross-entropy loss and set the batch size as 16.

\subsection{Baselines}
We adopt two types of baselines: 
\paragraph{Text-based Baselines} only encode text content: (1) PCNN \cite{zeng2015distant}, (2) BERT \cite{Devlin2019BERTPO}, and (3) MTB \cite{soares2019matching}.

\paragraph{Multi-modal Baselines}  encode both text and image contents: (1) UMT \cite{yu2020improving} adopts the multimodal interaction module to obtain the token representations incorporated with visual information and visual representations. (2) UMGF \cite{zhang2021multi} adopts a unified multi-modal graph fusion method. (3) BSG \cite{ZhengFFCL021} adopts the textual representation from BERT and the visual characteristics produced by the scene graph (SG). (4) MEGA \cite{ZhengWFF021} adopts a dual graph, which could align multi-modal features between entities and objects to improve performance. (5) VBERT \cite{li2019visualbert} adopts the single-stream structure which is different from the attention-based methods. (6) MoRe \cite{Wang2022NamedEA} obtains more textual information by retrieving images and titles, thereby improving the accuracy of relation classification and named entity recognition. (7) Iformer \cite{li2022analyzing} increases the amount of information in the image by detecting the objects. (8) HVPnet \cite{Chen2022} treats visual representations as visual prefixes that can be inserted to guide textual representations of error-insensitive prediction decisions.

\begin{table}
\centering
\scalebox{0.68}{
\begin{tabular}{cccccc}
\thickhline
\multicolumn{2}{c}{\multirow{1}{*}{Methods}} & Accuracy & Precision & Recall & F1  \\  
\midrule
\multirow{3}{*}{\makecell[c]{Text \\ Based}}  &
\multicolumn{1}{|l|}{PCNN} &  73.15 & 62.85 & 49.69 & 55.49 \\  
&\multicolumn{1}{|l|}{BERT} & 74.42  & 58.58 & 60.25 & 59.40 \\  
&\multicolumn{1}{|l|}{MTB} & 75.69  & 64.46 & 57.81 & 60.86 \\  
\midrule
\midrule
\multirow{8}{*}{\makecell[c]{Multi \\ modal}} & 
\multicolumn{1}{|l|}{UMT} & 77.84  & 62.93  &63.88 & 63.46  \\
 & 
\multicolumn{1}{|l|}{UMGF} & 79.27 & 64.38  & 66.23 & 65.29  \\
 & 
\multicolumn{1}{|l|}{BSG} & 77.15 & 62.95  & 62.65 & 62.80  \\
 & 
\multicolumn{1}{|l|}{MEGA} &  80.05 & 64.51  &68.44 &66.41 \\
 & 
\multicolumn{1}{|l|}{VBERT} & 73.97 & 57.15 & 59.48 & 58.30 \\
 & 
\multicolumn{1}{|l|}{MoRe} & 79.87   & 65.25 & 67.32 & 66.27 \\
 & 
\multicolumn{1}{|l|}{Iformer} & 92.38 & 82.59  & 80.78 &81.67  \\
 & 
\multicolumn{1}{|l|}{HVPnet} &  92.52 & 82.64  &80.78 &81.85 \\

\midrule
\midrule

\multicolumn{2}{l|}{\textbf{Ours}} &  \textbf{93.54\small±0.16} & \textbf{85.03\small±0.14}  &\textbf{84.25\small±0.17} &\textbf{84.64\small±0.16}  \\
\multicolumn{2}{l|}{\textit{w/o Object Evi.}} &  92.37\small±0.16 & 83.02\small±0.14 & 82.36\small±0.18 & 82.69\small±0.15 \\
\multicolumn{2}{l|}{\textit{w/o Image Evi.}} &  92.83\small±0.15 & 83.44\small±0.18 & 83.15\small±0.15 & 83.29\small±0.17 \\
\multicolumn{2}{l|}{\textit{w/o Visual Evi.}} &  92.72\small±0.17 & 82.78\small±0.19 & 83.63\small±0.24 &  83.20\small±0.21 \\
\multicolumn{2}{l|}{\textit{w/o Selection}} &  92.75\small±0.16 & 82.81\small±0.14 & 83.44\small±0.16 &  83.12\small±0.16 \\
\multicolumn{2}{l|}{\textit{w/o Consistency}} & 92.68\small±0.15 & 83.40\small±0.13  &82.71\small±0.16 & 83.05\small±0.15 \\

\thickhline
\end{tabular}}
\caption{The overall performance on MNRE.}
\label{tab:verification}
\vspace{-4mm}
\end{table}
\subsection{Main Results}
Table \ref{tab:verification} shows the mean and standard deviation results with 5 runs of training and testing on MRNE. We first compare text-based and multi-modal baselines and observe the performance improvement after incorporating visual content, indicating that images can help reveal the potential relationship between two entities. For the multi-modal model, Iformer~\cite{li2022analyzing} and HVPnet~\citep{Chen2022} specifically detect the objects in the image and achieve the average 17.23\% F1 and 14.15\% Accuracy compared with other multi-modal baselines. Therefore, we retrieve textual and visual evidence based on the object, sentence, and whole image, and achieve an average of 2.79\% F1 and 1.02\% Accuracy gains compared to the best-reported model HVPnet. Thanks to the retrieved visual and textual evidence, the text and image content in the original post is further explained, which helps our model obtain valuable clues to classify the relations between two entities.

\subsection{Analysis and Discussion}
\paragraph{Ablation Study.}
We conduct an ablation study to show the effectiveness of different modules of our model on the test set. Ours \textit{w/o Object Evidence} and Ours \textit{w/o Image Evidence} remove the descriptions of Objects and Images respectively in the retrieved textual evidence. Correspondingly, Ours \textit{w/o Visual Evidence} removes the visual evidence for text content retrieval. The results from Table \ref{tab:verification} demonstrate that the three types of evidence can bring 1.95\%, 1.35\%, and 1.44\% F1 improvements, respectively. Among them, the textual evidence obtained from the object retrieval brings the greatest benefit, which is also related to the potential entity information contained in the object. The removal of the \textit{Cross-Modal Selection} and \textit{Cross-Modal Consistency} modules means that we no longer use the appropriate evidence selection and update the retrieved evidence with the original content, which increases the noise from irrelevant evidence and leads to 1.52\% F1 and 1.59\% F1 down.

\paragraph{Analyze the Impact of Evidence.}

\begin{figure}
    \centering
    \includegraphics[scale=0.54]{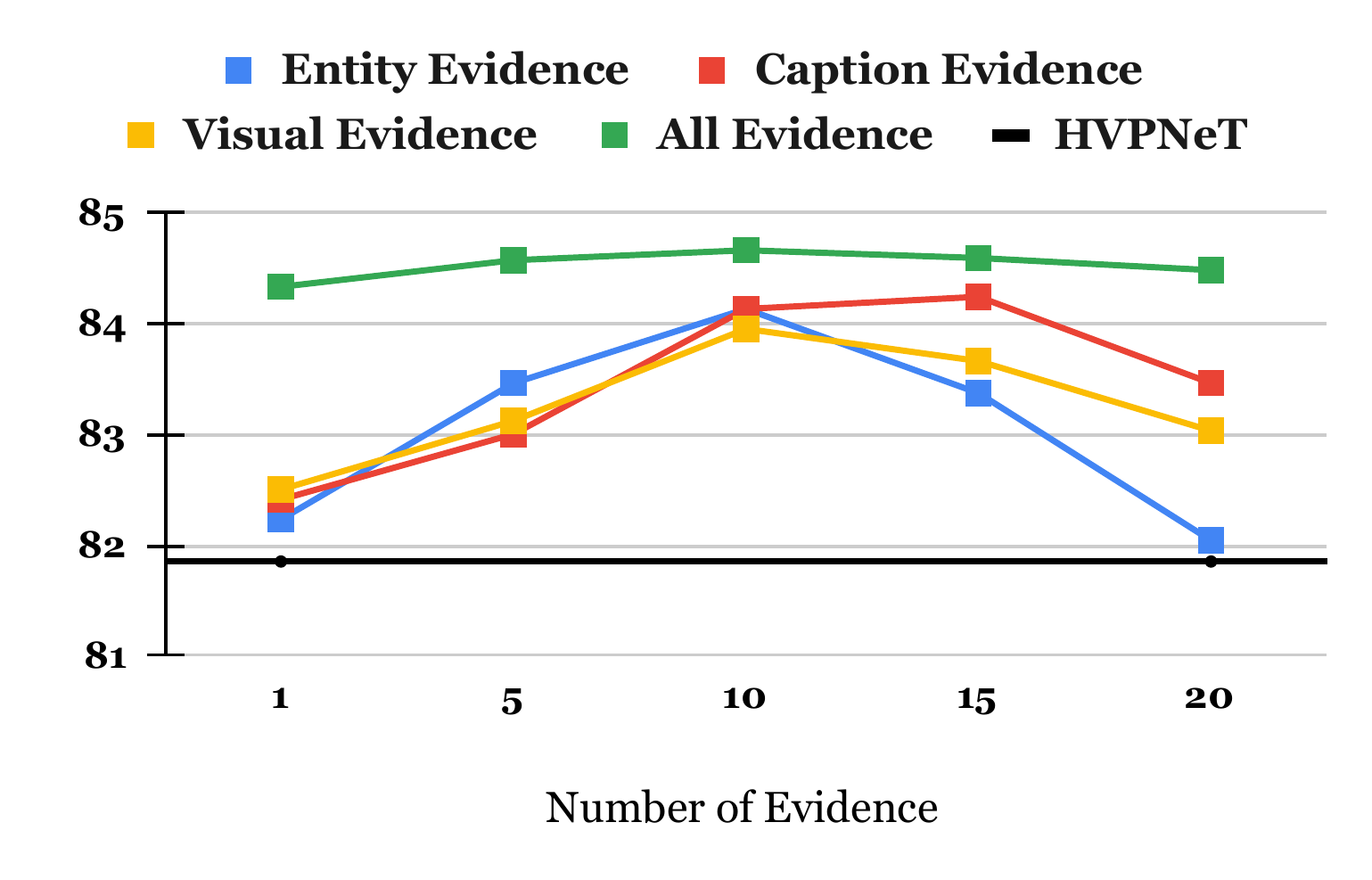}
    \caption{Comparison of different amounts of evidence.}
    \label{fig:evidence}
\end{figure}

In Figure~\ref{fig:evidence}, we vary the numbers of retrieved visual and textual evidence from $1\sim 20$ and report the F1 on the test set. The fluctuation results indicate that both the quantity and quality of retrieved evidence affect the performance. Using less textual or visual evidence cannot bring enough explanation to the original post, which leads to a decrease in the quality of the model classification. Using too much evidence will introduce false or irrelevant evidence noise, affecting performance. However, no matter how much evidence is adopted, our method consistently outperforms HVPnet, which illustrates the effectiveness of adding evidence. In our model, we adopt 10 textual and visual evidence for each post to achieve the best performance. We believe the Cross-Modal Consistency module can alleviate the irrelevant noise so that the model can obtain helpful auxiliary evidence.

\paragraph{Analyze Performance Changes in Tail Relations.}

\begin{figure}
    \centering
    \includegraphics[scale=0.42]{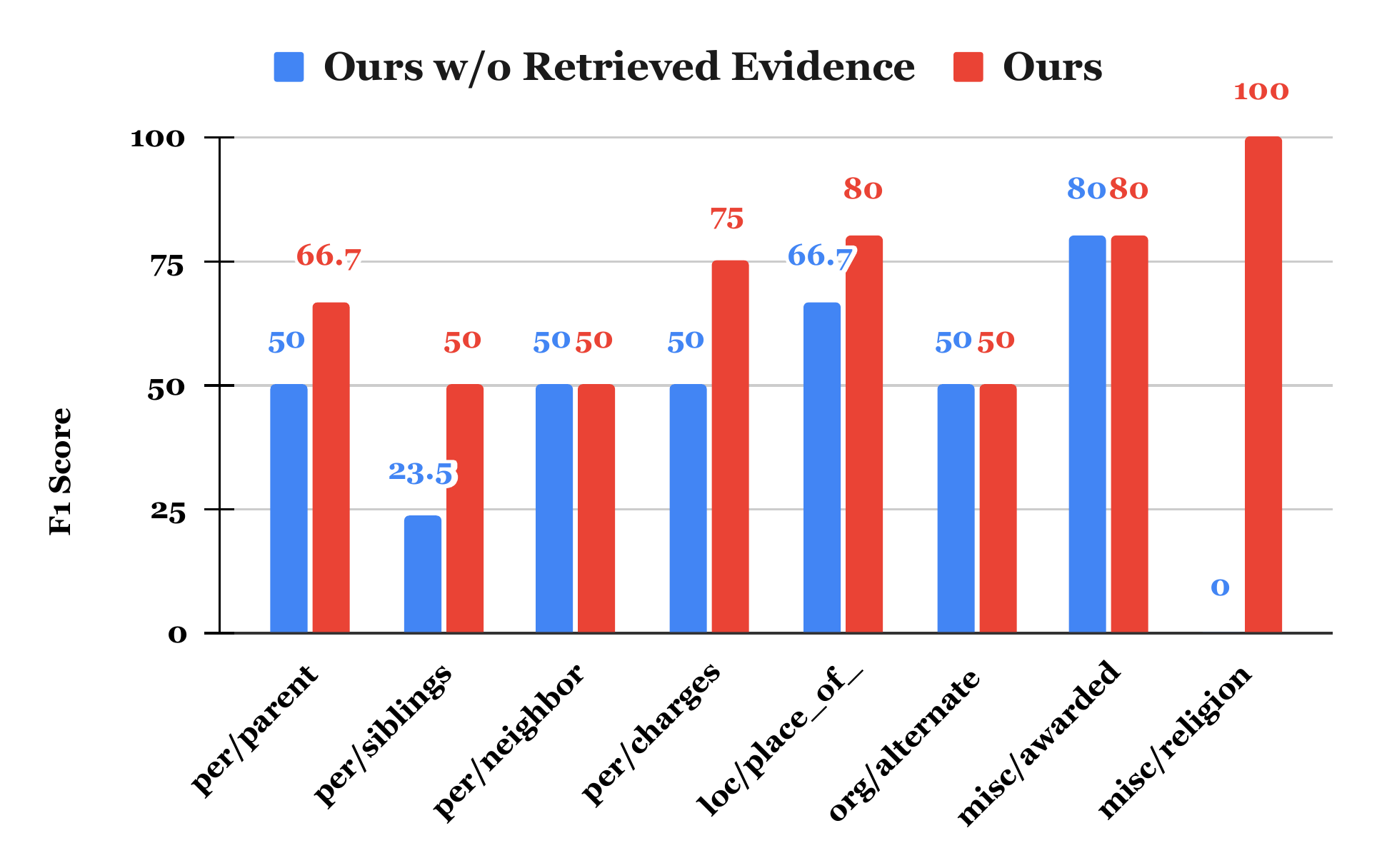}
    \caption{F1 performance changes of the tail relations.}
    \label{fig:change}
\end{figure}

We select the tail relations with the least number of data among the 23 relation classes in MNRE, and study their F1 performance changes after adding retrieval evidence in Figure \ref{fig:change}. Compared with the 2.79\% improvement brought by the evidence on all relations, we find that almost all tail relations can get more than 22.68\% F1 improvement (46.28 vs. 68.96), which shows that the retrieved evidence is more helpful for the few-shot tail relation types: It is an attractive property in real-world applications since classes of tail relations are usually more difficult to obtain training labeled data to improve.
\section{Related Work}
\label{sec:related_work}

Relation extraction has garnered considerable interest in the research community due to its essential role in various natural language processing applications~\citep{GuoZL19, NanGSL20, hu2021gradient, hu2021semi}. The initial efforts in this field focused on detecting relations between entities in the text, with different neural architectures~\citep{zeng2015distant,ZhangZCAM17,GuoN0C20} and pretrained language models~\citep{soares2019matching,Devlin2019BERTPO} used to encode the textual information. Multimodal relation extraction has recently been proposed, where visual clues from images are used to enhance entity representations~\citep{ZhengFFCL021,ZhengWFF021,Chen2022,Wang2022NamedEA}. Most existing efforts focus on fusing the visual and textual modalities effciently. \citet{ZhengWFF021} constructed the dual modality graph to align multi-modal features among entities and objects. \citet{Chen2022} concatenated object-level visual representation as the prefix of each self-attention layer in BERT. \citet{li2022analyzing} introduced a fine-grained multimodal fusion approach to align visual and textual objects in representation space. Closest to our work, \citet{Wang2022NamedEA} proposed to retrieve textual information related to the entities based on the given image and sentence. Unlike prior efforts, we not only retrieve texts related to entities but also retrieve visual and textual evidence related to the object, sentence, and entire image. We further synthesize the retrieved object-level, image-level, and sentence-level information for better reasoning between the same and different modalities. 
\section{Conclusion and Future Work}
\label{conclusion}

We propose to retrieve multimodal evidence and model the interactions among the object, sentence, and whole image for better relation extraction. Experiments show that the proposed method achieves competitive results on MNRE. For future research directions, we can utilize open-source image search and caption generation tools to retrieve textual and image evidence. For example, to retrieve visual evidence, one can (1) use a web crawler to search Google Images, or (2) utilize a searchable image database: PiGallery\footnote{\url{https://github.com/vladmandic/pigallery}}, where images can be sourced from Open Image Dataset\footnote{\url{https://storage. googleapis.com/openimages/web/factsfigures_v7.html}}, which contains $\sim$9 million images. For retrieving textual evidence, one can use CLIP to generate image captions. Moreover, we can also apply the method of multimodal retrieval to low-resource relation extraction \cite{hu2020selfore,liu2022hierarchical,hu2023think}, natural language inference \cite{li2023multi,li2022pair}, semantic parsing \cite{liu2022semantic,liu2023comprehensive}, and other NLP tasks, thus realizing information enhancement based on images and retrieval.

\section{Limitation}
\label{limitation}

In this paper, we suggest incorporating  textual and visual data from search engines for multimodal relation extraction. Despite the fact that the proposed model yields competitive results on the benchmark, it still has several limitations. Firstly, using a search engine is a feasible way to obtain related knowledge, but it also brings the issue of noisy evidence. Unrelated visual and textual evidence returned by the search engine may lead to incorrect predictions from the model. Additionally, not all the retrieved evidence is equally reliable, and sometimes sources may contradict each other. On the other hand, retrieval-augmented methods are slower than content-based counterparts, since retrieving evidence from the Internet requires extra time. Therefore, it may not satisfy some of the time-sensitive scenarios. Lastly, evidence may be presented in different forms other than texts and images. For instance, structural information such as tables, info lists, and knowledge graphs also provide important contexts for identifying semantic relations. Humans are able to extract relevant information from these heterogeneous sources for inference, while our relation extraction system can only model and reason over textual and visual evidence.

\section{Acknowledgement}
We thank the reviewers for their valuable comments. The work described here was partially supported by grants from the National Key Research and Development Program of China (No. 2018AAA0100204) and from the Research Grants Council of the Hong Kong Special Administrative Region, China (CUHK 14222922, RGC GRF, No. 2151185), NSF under grants III-1763325, III-1909323, III-2106758, and SaTC-1930941. Zhiyang Teng was partially supported by CAAI-Huawei MindSpore Open Fund (CAAIXSJLJJ-2021-046A).

\balance
\bibliography{custom}
\bibliographystyle{acl_natbib}


\end{document}